\documentclass[9pt,conference]{IEEEtran}
\usepackage{amssymb,amsthm,amsmath,array}
\usepackage{graphicx}
\usepackage[caption=false,font=footnotesize]{subfig}
\usepackage{xspace}
\usepackage[sort&compress, numbers]{natbib}
\usepackage{stmaryrd}
\usepackage{xcolor}
\usepackage{mathtools}
\usepackage{float}
\usepackage{textcomp}

\begin{document}
\title{Neuromorphic spintronics simulated using an unconventional data-driven Thiele equation approach}
\author{\IEEEauthorblockN{
        Anatole Moureaux\IEEEauthorrefmark{1}, 
        Simon de Wergifosse \IEEEauthorrefmark{1}, 
        Chloé Chopin\IEEEauthorrefmark{1}
        and 
        Flavio Abreu Araujo\IEEEauthorrefmark{1}
    }
    \IEEEauthorblockA{
        \IEEEauthorrefmark{1} Université catholique de Louvain}
}
\maketitle
\begin{abstract}
    In this study, we developed a quantitative description of the dynamics of spin-torque vortex nano-oscillators (STVOs) through an unconventional model based on the combination of the Thiele equation approach (TEA) and data from micromagnetic simulations (MMS). Solving the STVO dynamics with our analytical model allows to accelerate the simulations by 9 orders of magnitude compared to MMS while reaching the same level of accuracy. Here, we showcase our model by simulating a STVO-based neural network for solving a waveform classification task. We assess its performance with respect to the input signal current intensity and the level of noise that might affect such a system. Our approach is promising for accelerating the design of STVO-based neuromorphic computing devices while decreasing drastically its computational cost.
\end{abstract}

\section{Introduction}
The worldwide growth of artificial intelligence for the last decade is an indisputable phenomenon. Modern user-friendly applications such as the famous OpenAI's ChatGPT and Dall-E are nonetheless inseparable from a colossal energy consumption. As a result, the research for new devices reconciling cognitive performance and energy efficiency has become an expanding study field in the last few years. Neuromorphic computing addresses this problem by designing systems that mimic the architecture of the biological brain to reduce the energy required to learn a given task.

\begin{figure}[!ht]%
    \centering%
    \includegraphics[width=.4\textwidth]{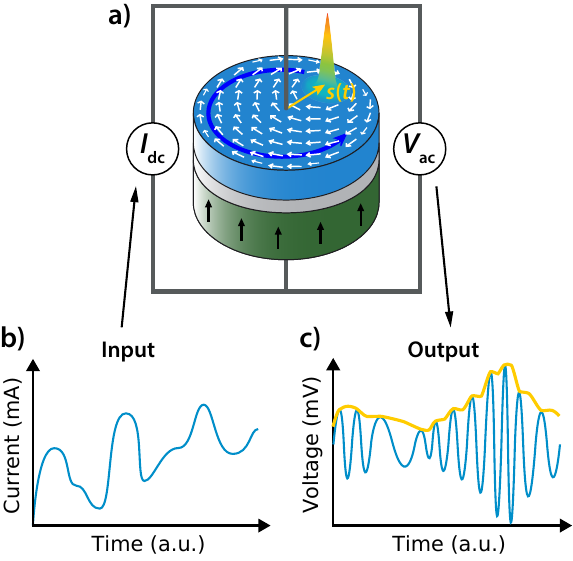}%
    \caption{a) The STVO-based system with an oscillating vortex core in the free magnetic layer (blue dot) resulting from the input dc current, b), the input signal, an amplitude-modulated dc current, and c), the output signal, an alternating voltage resulting from the dc  current excitation. Its envelope (in yellow) corresponds to the reduced position of the vortex core, and constitutes a non-linear transformation of the input signal.}%
    \label{fig:stvo}%
\end{figure}

For instance, hardware neural networks based on spin-torque vortex nano-oscillators (STVOs) proved to be promising implementations of neuromorphic computing\cite{larger2012photonic, torrejon2017neuromorphic, Romera2018, abreu2020role}. STVOs are magnetic tunnel junctions (see Fig. \ref{fig:stvo}a) playing the role of artificial neurons by processing the data with a complex non-linearity. STVO-based neural networks have already proved their ability to classify composite data such as spoken digits with a state-of-the-art accuracy level\cite{torrejon2017neuromorphic, abreu2020role}. The simulation of the STVO dynamics is of primary importance but none of the available state-of-the-art frameworks is however able to combine the accuracy and the speed of such simulations. Abreu Araujo \textit{et al.}\cite{araujo2022data} proposed a new framework based on the association of the Thiele equation approach (TEA) and micromagnetic simulations (MMS) called the \textit{data-driven Thiele equation approach} (DD-TEA), allowing to describe accurately the STVO dynamics over extended periods of time. %
In this study, we used DD-TEA to describe the STVO dynamics analytically. The resulting model allows to perform extended studies on the performance of a STVO-based neural network during a task of waveform classification. The influence of the input signal current intensity and the level of noise in the system on the quality of the recognition is then assessed.%

\section{Methods}
The ground state of the STVO free layer magnetization is a magnetic vortex whose core undergoes an in-plane gyrotropic oscillation when it is submitted to a sufficiently high electrical current intensity\cite{ralph2008spin, gaididei2010magnetic}. The reduced position $s$ of the vortex core depends non-linearly on the intensity of the injected current (Fig. \ref{fig:stvo}). One thus obtain a non-linear transformation of any amplitude-modulated input signal that is then readable in the output voltage of the structure thanks to the tunnel magnetoresistance effect.
By including the non-linear transient regime of the vortex core in the Thiele equation approach framework, it is possible to write 
\begin{equation}
    \dot{s}(t) = \alpha s(t) + \beta s^{n+1}(t)
    \label{eq:tea}
\end{equation}
where $\alpha$, $\beta$ and $n$ are parameters describing the non-linear STVO dynamics, and depend on the input signal current intensity and the geometry of the STVO. The value of these parameters can be accurately determined by fitting their analytical description with a small amount of MMS data. Equation (\ref{eq:tea}) is a Bernoulli differential equation and admits the following solution :
\begin{equation}
        s(t) = \dfrac{s_0}{\sqrt[n]{\left(1+\dfrac{s_0^n}{\alpha/\beta}\right)\exp\left(-n\alpha t\right)-\dfrac{s_0^n}{\alpha/\beta}}}
        \label{eq:hptea}
\end{equation}
where $s_0$ is the initial orbit of the vortex core. The use of MMS results in (\ref{eq:hptea}) allows to describe the STVO dynamics with a high level of accuracy and takes into account the full complexity of the vortex core dynamics. However, its analytical solving allows to speed up the simulations by $9$ orders of magnitude compared to full-MMS simulations. The time-dependent reduced position $s(t)$ of the vortex core then allows to effectively treat the data non-linearly. The data is thus projected in a space of much higher dimension, so that a further classification using a simple linear regression is made possible. %
A neural network composed of $24$ virtual neurons whose output was simulated using (\ref{eq:hptea}) was created. It was then set to classify sine and square signals in the framework of reservoir computing\cite{paquot2010reservoir, larger2012photonic, torrejon2017neuromorphic, abreu2020role}. The accuracy and the root-mean-square error (RMSE) between the expected and the actual outputs were computed in order to investigate the performance of the neural network under specific operating conditions. The process was repeated over a sweep of the input signal current intensity and the level of noise in the system to assess their influence on the quality of the recognition. Each data point in the sweep (see Fig. \ref{fig:param})
corresponds to the average over $200$ simulations to properly take into account the random fluctuations introduced by the noise. The procedure is discussed with more details in our recent manuscript\cite{moureaux2023}.%
\section{Results}%
\begin{figure}
    \centering
    \includegraphics[width=.5\textwidth, trim={.5cm 0cm .5cm 0cm}, clip]{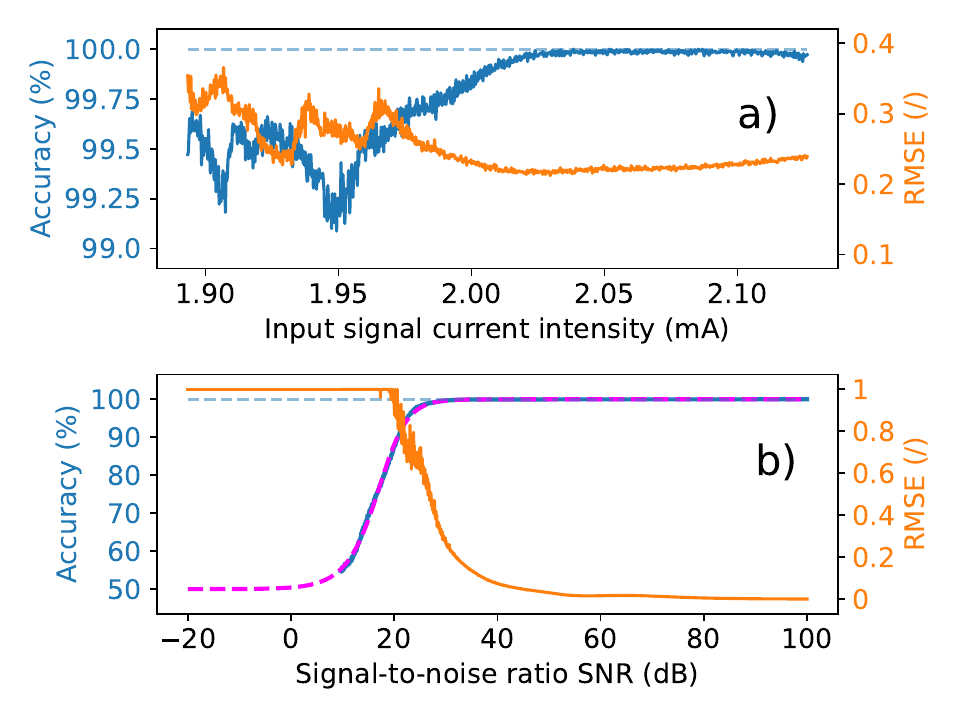}
    \caption{Accuracy and RMSE of the STVO-based neural network during sine/square classification with respect to a) the intensity of the input signal current and b) the signal-to-noise ratio. The pink curve is a generalized logistic fit. 
    }
    \label{fig:param}
\end{figure}

Here, an improvement of the recognition rate performance at high working currents (see Fig. \ref{fig:param}a) is clearly shown. This implies that the dynamics of the STVO can be operated at an optimal current. Indeed, the STVO dynamics exhibits a higher non-linear behaviour at higher currents, hence allowing a better quality of data processing\cite{abreu2020role}. For lower currents on the contrary, a degradation is shown due to the lower probability of the signal to generate oscillations that explore enough non-linearity to treat the input data. %
A drastic improvement of the recognition quality at higher signal-to-noise ratios (SNRs) can also be seen in Fig. \ref{fig:param}b due to the cleaner STVO dynamics. Under $0$ dB, the accuracy falls down to $50\%$ as the noise starts to be more important than the input signal. The neural network fails to detect usable features in the data and yields a random classification between the sine and square signals. The RMSE explodes for lower SNRs due to the detrimental influence of the noise on the behavior of the STVOs. %
These studies, which were not feasible with former simulations techniques such as full-MMS due to their heavy required execution time and computing power, enable the determination of high added value information about STVO-based neural networks like the minimum signal current intensity or the maximum level of noise admitted to reach a given level of accuracy. Our model has the potential to predict and accelerate experimental investigations by decreasing their computational cost, and it is expected to allow the investigation of much more complex architectures related to neuromorphic computing applications.%
\section*{Acknowledgement}
Computational resources have been provided by the Consortium des Équipements de Calcul Intensif (CÉCI), funded by the 
F.R.S.-FNRS under Grant No. 2.5020.11 and by the Walloon Region. 
F.A.A. is a Research Associate and S.d.W. is a FRIA grantee, both of the F.R.S.-FNRS.
\bibliographystyle{IEEEtran}
\bibliography{references}
\end{document}